%% file: main.tex
\pdfoutput=1

\documentclass[11pt]{article}

\usepackage{acl}

\usepackage{times,todonotes}
\usepackage{latexsym}

\usepackage[T1]{fontenc}

\usepackage[utf8]{inputenc}

\usepackage{microtype}

\usepackage{defs}

\usepackage[ruled,vlined,linesnumbered]{algorithm2e}
\usepackage{adjustbox}

\usepackage{forest} %
\usepackage{caption}
\usepackage{subcaption}
\usepackage{float}
\usepackage{placeins}
\usepackage{inconsolata}

\newcommand{\sysname}{\textsc{ReFill}} %
\newcommand{\ratsql}{\textsc{RAT}}

\newcommand{\bert}{\textsc{BERT}}
\newcommand{\bart}{\textsc{BART}}

\newcommand{\smbop}{\textsc{SmBoP}}

\newcommand{\gazp}{\textsc{GAZP}}
\newcommand{\ltos}{\textsc{L2S}}
\newcommand{\snowball}{\textsc{SnowBall}}
\newcommand{\baseline}{\textsc{Base-M}}

\newcommand{\texttosql}{Text-to-SQL}
\newcommand{\sqltotext}{SQL-to-Text}
\newcommand{\nlqs}{{\mathcal{X}}}

\newcommand{\schemas}{{\mathcal{S}}}
\newcommand{\trainschemas}{{\schemas_\text{train}}}

\newcommand{\queries}{{\mathcal{Q}}}
\newcommand{\model}{{\mathcal{M}}}
\newcommand{\textgen}{\mathcal{B}}
\newcommand{\Filter}{{\mathcal{F}}}

\newcommand{\trainset}{{\mathcal{D}_\text{train}}}

\newcommand{\vxm}{\vx^\text{masked}}
\newcommand{\vqe}{\vq^\text{Eng}}
\newcommand{\mask}{{\tt MASK}}
\newcommand{\workload}{{\mathcal{QW}_s}}
\newcommand{\dsyn}{\mathcal{D}_\text{syn}}

\title{Diverse Parallel Data Synthesis for Cross-Database Adaptation of Text-to-SQL Parsers}

\newcommand{\aspace}{\hspace{1.0em}}

\author{Abhijeet Awasthi \aspace 
        Ashutosh Sathe \aspace 
        Sunita Sarawagi \aspace 
        \\
        Indian Institute of Technology Bombay, India \\
        \texttt{\{awasthi,absathe,sunita\}@cse.iitb.ac.in}
        }

\begin{document}
\maketitle
\begin{abstract}
Text-to-SQL parsers typically struggle with databases unseen during the train time. Adapting parsers to new databases is a challenging problem due to the lack of natural language queries in the new schemas. We present \sysname{}, a framework for synthesizing high-quality and textually diverse parallel datasets for adapting a Text-to-SQL parser to a target schema. \sysname{} learns to retrieve-and-edit text queries from the existing schemas and transfers them to the target schema. We show that retrieving diverse existing text, masking their schema-specific tokens, and refilling with tokens relevant to the target schema, leads to significantly more diverse text queries than achievable by standard SQL-to-Text generation methods. Through experiments spanning multiple databases, we demonstrate that fine-tuning parsers on datasets synthesized using \sysname{} consistently outperforms the prior data-augmentation methods.    
\end{abstract}

\section{Introduction}
Natural Language interface to Databases (NLIDB) that translate text queries to executable SQLs is a challenging task in the field of Semantic Parsing~\cite{geoquery-original, zettlemoyerlearning, berant2013semantic}. In addition to understanding the natural language and generating an executable output, Text-to-SQL also requires the ability to reason over the schema structure of relational databases.  Recently, datasets such as Spider~\cite{spider2018yu} comprising of parallel (Text,SQL) pairs over hundreds of schemas have been released, and these have been used to train state-of-the-art neural \texttosql\ models~\cite{ratsql2020wang,duorat2021scholak,smbop2021rubin,picardScholak2021,dtfixup2021xu}.  However, several studies have independently shown that such {\texttosql} models fail catastrophically when evaluated on unseen schemas from the real-world databases~\cite{suhr2020exploring,kaggledbqa2021lee,wildtext2sql2021hazoom}.  Adapting existing parsers to new schemas is challenging due to the lack of parallel data for fine-tuning the parser. 

Synthesizing parallel data, that is representative of natural human generated queries~\cite{overnight2015,herzig2019don}, is a long-standing problem in semantic parsing. Several methods have been proposed for supplementing with synthetic data, ranging from grammar-based canonical queries to full-fledged conditional text generation models~\cite{overnight2015, herzig2019don, grounded-2020-zhang, yang2021hierarchical, hierarchical-zhang-2021, victorialin2021}. For \texttosql, data-augmentation methods are primarily based on training an \sqltotext\ model using labeled data from pre-existing schemas, and generating data in the new schemas. 
We show that the text generated by these methods, while more natural than canonical queries, lacks the rich diversity of natural multi-user queries. Fine-tuning with such data often deteriorates the model performance since the lack of diversity leads to a biased model.

We propose a framework called \sysname{} (\S~\ref{sec:our_method}) for generating diverse text queries for a given SQL workload that is often readily available~\cite{BaikJ019}. \sysname\ leverages parallel datasets from several existing schemas, such as Spider~\cite{spider2018yu}, to first retrieve a diverse set of text paired with SQLs that are structurally similar to a given SQL $\vq$ (\S~\ref{sec:retrieve_queries}). Then, it trains a novel {\em schema translator} model for converting the text of the training schema to the target schema of $\vq$.  The schema translator is decomposed into a {\tt mask} and {\tt fill} step to facilitate training without direct parallel examples of schema translation.  Our design of the {\tt mask} module and our method of creating labeled data for the {\tt fill} module entails non-trivial details that we explain in this paper (\S~\ref{sec:schema_translation}). \sysname\ also incorporates a method of filtering-out inconsistent (Text,SQL) pairs using an independent binary classifier (\S~\ref{sec:filter_desc}), that provides more useful quality scores, than the cycle-consistency based filtering~\cite{grounded-2020-zhang}.  Our approach is related to retrieve-and-edit models that have been used for semantic parsing~\cite{hashimoto2018retrieve}, dialogue generation~\cite{chi2021neural}, translation~\cite{cai-2021-nmt}, and question answering~\cite{karpukhin2020dense}.  However, our method of casting the "edit" as a two-step mask-and-fill schema translation model is different from the prior work.

We summarize our contributions as follows:
(i)~We propose the idea of retrieving and editing natural text from several existing schemas for transferring it to a target schema, obtaining higher text diversity compared to the standard {\sqltotext} generators.
(ii)~We design strategies for masking schema-specific words in the retrieved text and training the {\sysname} model to fill in the masked positions with words relevant to the target schema. 
(iii)~We filter high-quality parallel data using a binary classifier and show that it is more efficient than existing methods based on cycle-consistency filtering.
(iv)~We compare {\sysname} with prior data-augmentation methods across multiple schemas and consistently observe that fine-tuning {\texttosql} parsers on data generated by {\sysname} leads to more accurate adaptation.

\section{Diverse data synthesis with {\sysname}}
\label{sec:our_method}
Our goal is to generate synthetic parallel data to adapt an existing \texttosql{} model to a target schema unseen during training.  A {\texttosql} model  $\model\!:\!\nlqs,\schemas\!\mapsto\!\queries$   maps a natural language question $\vx \in \nlqs$ for a database schema $\vs \in \schemas$,
to an SQL query $\hat{\vq} \in \queries$.   We assume a {\texttosql} model $\model$ trained on a dataset $\trainset = \{(\vx_i, \vs_i, \vq_i)\}_{i=1}^{N}$ consisting of text queries $\vx_i$ for a database schema $\vs_i$, and the corresponding gold SQLs $\vq_i$.  %
The train set $\trainset$ typically consists of examples from a wide range of schemas $\vs_i \in \trainschemas $. For example, the Spider dataset~\cite{spider2018yu} contains roughly 140 schemas in the train set. %
We focus on adapting the model $\model$ to perform well on a target schema $\vs$ different from the training schemas in $\trainschemas$. %
To achieve this, we present a method of generating synthetic data $\dsyn$ of Text-SQL pairs containing diverse text queries for the target schema $\vs$. We fine-tune the model $\model$ on $\dsyn$ to adapt it to the schema $\vs$. Our method is agnostic to the exact model used for \texttosql{} parsing. %
We assume that on the new schema $\vs$ we have a workload $\workload$ of SQL queries. Often in existing databases
a substantial SQL workload is already available in the query logs at the point a DB manager decides to incorporate the NL querying capabilities~\cite{BaikJ019}.  The workload is assumed to be representative but not exhaustive.  In the absence of a real workload, a grammar-based SQL generator may be used  %
~\cite{grounded-2020-zhang, victorialin2021}. 

\begin{figure*}[h!]
\centering
\includegraphics[width=\textwidth]{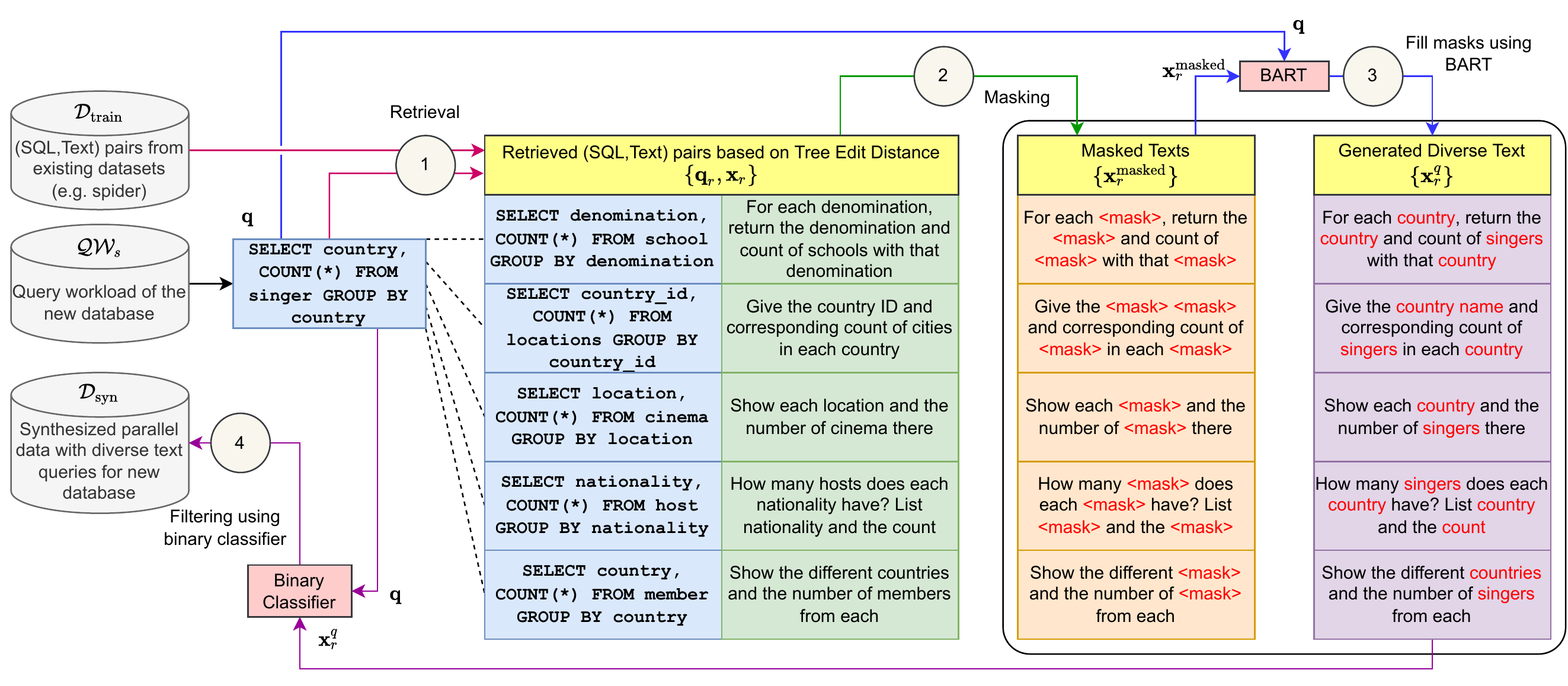}
\caption{Diverse parallel data synthesis by retrieving-and-editing existing examples using {\sysname}. Given an SQL $\vq$ from a new schema, {\sysname} \textbf{(1)}~\underline{Re}trieves SQL-Text pairs from an existing dataset (\S~\ref{sec:retrieve_queries}) where the SQLs are structurally similar to $\vq$ (indicated by dashed lines). \textbf{(2)}~Since the retrieved text come from a different schema, we mask-out the schema-specific words (\S~\ref{sec:schema_translation}). \textbf{(3)} The masked text and the SQL $\vq$ are then translated into the target schema via an Edit and \underline{Fill} step that uses a conditional text generation model like BART (\S~\ref{sec:schema_translation}). In this way, we transfer the text from multiple existing schemas to generate diverse text for the new schemas.
\textbf{(4)}~Finally, we use a binary classifier as a filtering model to retain only the consistent Text-SQL pairs in the output dataset (\S~\ref{sec:filter_desc}). }
\label{fig:main_figure}
\end{figure*}

\begin{algorithm}[t]
\small
\textbf{input:} $\workload$, $\model$, $\trainset$ \\
$\dsyn \leftarrow \phi $ \\
\For{$\vq \leftarrow$ $\operatorname{SampleSQLQueries}$ ($\workload$)}
{
    $\{\vq_r, \vx_r\} \leftarrow \textbf{RetrieveRelatedPairs}(\vq, \trainset) $ \\
    $\{\vxm_r\} \leftarrow \textbf{MaskSchemaTokens}(\{\vq_r, \vx_r\}) $ \\
    $\{\vx_r^q\} \leftarrow \textbf{EditAndFill}(\{\vq, \vxm_r\}) $ \\
    $\dsyn \leftarrow \dsyn \cup \textbf{Filter}(\vq, \{\vx_r^q\}) $  \\
}
    $\model_\text{new} \leftarrow \text{fine-tune}(\model, \dsyn)$ %
\caption{Data Synthesis with {\sysname}}
\label{alg:highlevel}
\end{algorithm}

Figure~\ref{fig:main_figure} and Algorithm~\ref{alg:highlevel} summarizes our method for converting a workload $\workload$ of SQL queries into a synthetic dataset $\dsyn$ of Text-SQL pairs containing diverse text queries. %
Given an  SQL query $\vq \in \workload$  for the target schema $\vs$, our method first retrieves related SQL-Text pairs $\{\vq_r, \vx_r\}_{r=1}^{R}$
from $\trainset$ on the basis of a tree-edit-distance measure such that the SQLs $\{\vq_r\}_{r=1}^R$ in the retrieved pairs are structurally similar to the SQL $\vq$ (\S~\ref{sec:retrieve_queries}).  We then translate each retrieved text query $\vx_r$ so that its target SQL changes from $\vq_r$ to $\vq$ on schema $\vs$ (\S~\ref{sec:schema_translation}).  We decompose this task into two steps:
masking out schema specific tokens in $\vx_r$, and filling the masked text to make it consistent with $\vq$ using a conditional text generation model $\textgen$ like BART~\cite{lewis2020bart}.  The translated text may be noisy since we do not have direct supervision to train such models. Thus, to improve the overall quality of the synthesized data we filter out the inconsistent SQL-Text pairs using an independent binary classifier (\S~\ref{sec:filter_desc}). 
Finally, we adapt the {\texttosql} model $\model$ for the target schema $\vs$ by fine-tuning it on the diverse, high-quality filtered data $\dsyn$ synthesized by {\sysname}. 

\subsection{Retrieving related queries}
\label{sec:retrieve_queries}
\begin{figure}[t]
\centering
\includegraphics[width=0.45\textwidth,height=1.6in]{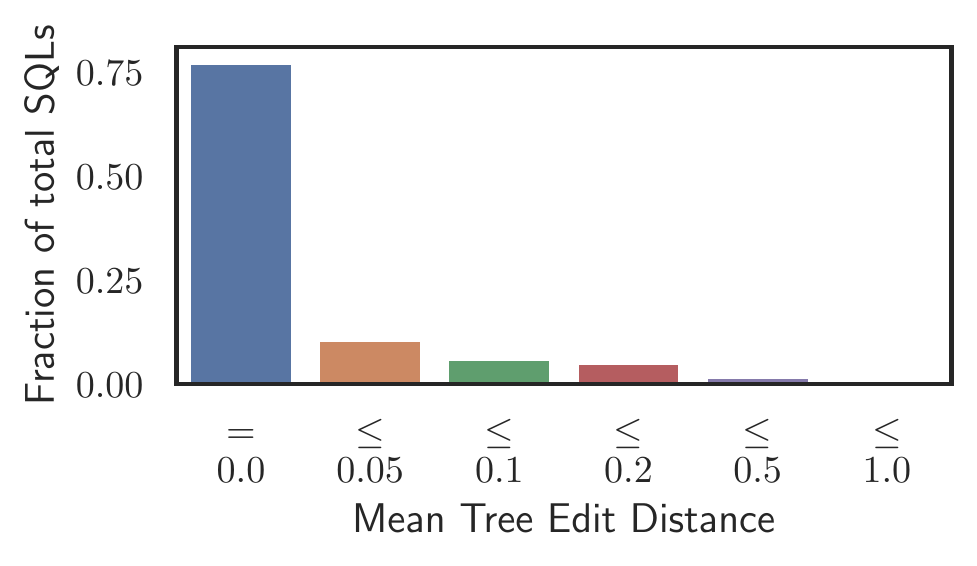}
\caption{Frequency distribution of average tree-edit-distance between SQLs and their three nearest neighbours from other schemas within Spider's train set.
}
\label{fig:teds}
\end{figure}

Given an SQL $\vq \in \workload$ sampled from SQL workload, we extract SQL-Text pairs $\{\vq_r, \vx_r\} \in \trainset $, from the train set such that the retrieved SQLs $\{\vq_r\}$ are structurally similar to the SQL $\vq$. We utilize tree-edit-distance~\cite{pawlik2015efficient,pawlik2016tree} between the relational algebra trees of SQLs $\vq$ and $\vq_r$ --- smaller distance implies higher structural similarity. Since the retrieved SQLs come from different schemas, we modify the tree-edit-distance algorithm to ignore the schema names and the database values. The tree-edit-distance is further normalized by the size of the larger tree. We only consider the $\{\vq_r, \vx_r\}$ pairs where the SQLs $\{\vq_r\}$ have a distance of less than $0.1$ w.r.t. the SQL $\vq$. Within datasets like Spider that span hundreds of schemas, it is often possible to find several SQLs structurally similar to a given SQL $\vq$.  For example, in Spider we found that 76\% of the train SQLs contain at least three zero-distance (structurally identical) neighbours in other schemas.  In Figure~\ref{fig:teds}, we present more detailed statistics.

\subsection{Translating text of related queries}
\label{sec:schema_translation}
Our next goal is to translate the retrieved $\vx_r$ from being a text for SQL $\vq_r$ %
to a text $\hat{\vx}$ for SQL $\vq$ %
, where $\vq\approx \vq_r$ structurally. %
However, we do not have a readily labeled dataset to learn a model that translates $\vx_r$ to $\hat{\vx}$ while being consistent with $\vq$.  
We therefore decompose this task into two steps: 1)~A simpler task of masking schema-specific tokens in $\vx_r$ to get a template $\vxm_r$ and 2)~A conditional text generation model that maps $(\vxm_r,\vq)$ to the text $\hat{\vx}$ consistent with $\vq$, by filling the masked positions in $\vxm_r$ as per $\vq$. We re-purpose $\trainset$ to get indirect supervision for training the text generation model. We now present each step in detail.

\paragraph{Masking the retrieved text}
Converting the retrieved text queries $\{\vx_r\}$ to masked templates $\{\vxm_r\}$ is a critical component of {\sysname}'s pipeline since irrelevant tokens like references to schema elements of the original database can potentially misguide the text generation module.  Our initial approach was to mask tokens based on a %
match of text tokens with schema names and  manually refined schema-to-text linked annotations as in~\citet{lei-etal-2020-examining}.  However, this approach failed to mask all schema-related terms since their occurrences in natural text often differed significantly from schema names in the database. Table~\ref{tab:mask_anecdotes} shows some anecdotes. Consequently, we designed a simple frequency-based method of masking that is significantly more effective for our goal of using the masked text to just guide the diversity.
For each word that appears in the text queries of the train set, we count the number of distinct databases where that word gets mentioned at least once. For example, common words like {\tt \{`show', `what', `list', `order'\}} get mentioned in more than 90\% of the schemas, and domain specific words like {\tt \{`countries', `government'\}} occur only in text queries of a few schemas. 
We mask out all the words that appear in less than 50\% of the schemas. The words to be masked are replaced by a special token {\mask}, and consecutive occurrences of {\mask} are collapsed into a single  {\mask} token. Thus we obtain masked templates $\{\vxm_r\}$ retaining minimal information about their original schema.

\paragraph{Editing and Filling the masked text}
Given a masked template $\vxm_r$, and an SQL query $\vq$, we wish to edit and fill the masked portions in $\vxm_r$ to make it consistent with the SQL $\vq$.
We utilize a conditional text generation model $\textgen$ like {\bart}~\cite{lewis2020bart} for this purpose.
We first convert $\vq$ into a pseudo-English representation $\vqe$ similar to~\citet{shu2021logic}, to make it easier for $\textgen$ to encode $\vq$. In addition, we wrap the table, column, or value tokens in $\vqe$ with special tokens to provide explicit signals to the text generation model $\textgen$ that such tokens are likely to appear in the output text $\hat{\vx}$. Next, we concatenate the tokens in %
$\vxm_r$ and $\vqe$ for jointly encoding them as an input to $\textgen$. The output of $\textgen$'s decoder is text $\hat{\vx}$, which is expected to be consistent with the SQL $\vq$. 

Since we do not have direct supervision to fine-tune $\textgen$ for this task, we present a method of repurposing $\trainset$ for fine-tuning $\textgen$. $\trainset$ contains SQL-Text pairs $(\vq_i, \vx_i)$ from various schemas $\vs_i$. A {\em Na\"ive} way to train $\textgen$ is to provide $[\vxm_i|\vqe_i]$, the concatenation of $\vxm_i$ and $\vqe_i$ as an input to the encoder and maximize the likelihood of $\vx_i$ in the decoder's output. This way the decoder of $\textgen$ learns to refill the masked tokens in $\vxm_i$ by attending to $\vqe_i$ to recover $\vx_i$ in the output. While useful for learning to refill the masked positions, this {\em Na\"ive} method of training $\textgen$ is mismatched from its use during inference in two ways: (i)~For a given SQL $\vq$, {\sysname} might fail to retrieve a similar structure neighbour of $\vq_i$ from $\trainset$. In such cases, $\textgen$ should be capable of falling back to pure {\sqltotext} generation mode to directly translate $\vq$ into $\hat{\vx}$. %
(ii)~During inference, $\vxm_r$ and $\vq$ come from different schemas. However, during {\em Na\"ive} training, the masked text $\vxm_i$ and the SQL $\vq_i$ are derived from the same example $(\vq_i, \vx_i)$.  To address these two limitations, we train $\textgen$ in a more {\em Robust} manner as follows:
(a)~For a random one-third of the train steps we train $\textgen$ in the Na\"ive way, allowing $\textgen$ to learn the filling of the masked tokens using $\vqe_i$. %
(b)~For another one-third, we pass only $\vqe_i$ as an input and maximize the likelihood of $\vx_i$. This ensures that model is capable of generating the text from the $\vqe_i$ alone, if the templates $\vxm_i$ are unavailable or noisy. (c)~For the remaining one-third, we first retrieve an SQL-Text pair $(\vq_j, \vx_j)$, from a different schema such that the SQL $\vq_j$ is structurally similar to $\vq_i$ (\S~\ref{sec:retrieve_queries}), and the word edit distance between the masked templates $\vxm_i$ and $\vxm_j$ is also small. We can then replace $\vxm_i$ with $\vxm_j$ and encode $[\vxm_j|\vqe_i]$ as an input to $\textgen$ and maximize the likelihood of $\vx_i$ in the decoder's output. This step makes the training more consistent with the inference, as $\vxm_j$ and $\vqe_i$ now come from different schemas.
In \S~\ref{sec:designchoice}, we justify training {\em Robustly} compared to {\em Na\"ive} training. %

\subsection{Filtering the Generated Text}
\label{sec:filter_desc}
Since the data synthesized using {\sysname} is used to fine-tune a downstream {\texttosql} parser, we  learn a Filtering model $\Filter: (\nlqs , \queries) \mapsto \mathbb{R}$ to discard inconsistent examples from the generated dataset. $\Filter$ assigns lower scores to inconsistent Text-SQL pairs. For each SQL $\vq \in \workload$, we select the top-5 sentences generated by {\sysname} and discard all the sentences that are scored below a fixed threshold as per the filtering model. 
Existing work depended on a trained \texttosql{} parser $\model$ to assign cycle-consistency scores~\cite{grounded-2020-zhang}. However, we show that cycle-consistency filtering favors text on which $\model$ already performs well, and hence does not result in a useful dataset for fine-tuning $\model$.

We instead train a filtering model $\Filter$ as a binary classifier, independent of $\model$. The Text-SQL pairs $\{(\vx_i, \vq_i)\}$ in the training set $\trainset$, serve as positive (consistent) examples and we synthetically generate the negative (inconsistent) examples as follows: (i)~Replace DB values in the SQL $\vq_i$ with arbitrary values sampled from the same column of the database. (ii)~Replace SQL-specific tokens in $\vq_i$ with their corresponding alternates e.g. replace \texttt{ASC} with \texttt{DESC}, or `$>$' with `$<$'. (iii)~Cascade previous two perturbations. (iv)~Replace the entire SQL $\vq_i$ with a randomly chosen SQL $\vq_j$ from the same schema. (v)~Randomly drop tokens in the text query $\vx_i$ with a fixed probability of 0.3. (vi)~Shuffle a span of tokens in the text query $\vx_i$, with span length set to 30\% of the length of $\vx_i$. Thus, for a given Text-SQL pair $(\vx_i, \vq_i)$ we obtain six corresponding negative pairs $\{(\vx^n_j, \vq^n_j)\}_{j=1}^{6}$. Let $s_i$ be the score provided by the filtering model for the original pair $(\vx_i, \vq_i)$ and $\{s_j\}_{j=1}^6$ be the scores assigned to the corresponding negative pairs $\{(\vx^n_j, \vq^n_j)\}_{j=1}^{6}$. We supervise the scores from the filtering model using a binary-cross-entropy loss over the Sigmoid activations of scores as in Equation~\ref{eq:bce}. 
\begin{equation}
    \mathcal{L}_\text{bce} = -\log\sigma(s_i) - \sum_{j=1}^{6}\log\sigma(1-s_j)
    \label{eq:bce}
\end{equation}
To explicitly contrast an original pair with its corresponding negative pairs we further add another Softmax-Cross-Entropy loss term.
\begin{equation}
    \mathcal{L}_\text{xent} = - \log\frac{\exp(s_i)}{\exp(s_i)+\sum_{j=1}^{6}\exp(s_j)}
    \label{eq:xentbc}
\end{equation}

\section{Related Work}
\label{sec:related_work}
\paragraph{SQL-to-Text generation}
Many prior works perform training data augmentation via pre-trained text generation models %
that translate SQLs into natural text~\cite{guo2018question, grounded-2020-zhang,gap-shi-2020,hierarchical-zhang-2021, victorialin2021, yang2021hierarchical, shu2021logic}. For example, ~\citet{victorialin2021} fine-tune BART~\cite{lewis2020bart} on parallel SQL-Text pairs to learn an {\sqltotext} translation model. ~\citet{shu2021logic} propose a similar model that is trained in an iterative-adversarial way along with an evaluator model. The evaluator learns to identify inconsistent SQL-Text pairs, similar to our filtering model. %
To retain high quality synthesized data~\citet{grounded-2020-zhang} additionally filter out the synthesized pairs using a pre-trained {\texttosql} model based on cycle consistency, that we show to be sub-optimal~(\S~\ref{sec:filtering}). The SQL workload in most of the prior work was typically sampled from hand-crafted templates or a grammar like PCFG induced from existing SQLs, or crawling SQLs from open-source repositories~\citet{gap-shi-2020}. However, database practitioners have recently drawn attention to the fact that SQL workloads are often pre-existing and should be utilized~\cite{BaikJ019}. %
\nocite{Wolfson2021WeaklySM}

\paragraph{Retrieve and Edit Methods}
Our method is related to the retrieve-and-edit framework, which has been previously applied in various NLP tasks. In Semantic Parsing, question and logical-form pairs from the training data relevant to the test-input question are retrieved and edited to generate the output logical forms in different ways~\cite{relationatt2018, cbr2021das, googlecbr2021, gupta2021retronlu}. In machine translation, memory augmentation methods retrieve-and-edit examples from translation memory to guide the decoder's output~\cite{hossain2020simple, cai-2021-nmt}. Our editing step --- masking followed by refilling is similar to style transfer methods that minimally modify the input sentence with help of retrieved examples corresponding to a target attribute~\cite{li2018delete}. In contrast to learning a retriever, we find simple tree-edit distance to be an effective metric for retrieving the relevant examples for our task. %

\input{tables/main_results}
\section{Experimental Set-up\footnote{~Code: \href{https://github.com/awasthiabhijeet/refill}{github.com/awasthiabhijeet/refill}}}
\label{sec:adapt}
We adapt pretrained {\texttosql} parsers on multiple database schemas unseen during the train time. Here, we describe the datasets, models, and evaluation metrics used in our experiments.

\noindent
{\bf Datasets}: We primarily experiment with the Spider dataset~\cite{spider2018yu}. Spider's train split contains 7000 {\texttosql} examples spanning 140 database schemas, and the dev split contains roughly 1000 examples spanning 20 schemas\footnote{Spider's test-split is inaccessible as of 10/26/2022}. Since individual schemas in the dev split typically contain less than 50 examples, to evaluate on a larger set of examples we adapt and evaluate the {\texttosql} parser on groups of similar schemas instead of individual schemas. We create 4 groups, with each group having database schemas from a similar topic. For example, Group-1 consists of databases {\tt \{Singer, Orchestra, Concerts\}}. We utilize all the available Text-SQL pairs in each group for evaluation. In appendix Table~\ref{sec:grouping_details}, we provide detailed statistics about each group. On average, each group contains 69 unique SQLs and 131 evaluation examples. To simulate a query workload $\workload$ for each group, we randomly select 70\% of the available SQLs and replace the constant-values in the SQLs with values sampled from their corresponding column in the database. We also evaluate on query workloads of size 30\% and 50\% of the available SQL queries. The SQL queries in the workload are translated using {\sysname} or an {\sqltotext} model, and the resulting Text-SQL pairs are then used to fine-tune a base {\texttosql} parser.

We further experiment with four datasets outside Spider in Section~\ref{sec:additional_datasets}. We work with GeoQuery~\cite{geoquery-original}, Academic~\cite{data-academic}, IMDB and Yelp~\cite{data-sql-imdb-yelp}. We utilize the pre-processed version of these datasets open-sourced by~\citet{spider2018yu}. In appendix Table~\ref{tab:non_spider_details}, we present statistics about each of the four datasets.

\noindent
{\bf {\texttosql} parser}: We experiment with  {\smbop}~\cite{smbop2021rubin} as our base {\texttosql} parser, and utilize author's implementation.  The {\smbop} model is initialized with a \textsc{RoBERTa-base} model, followed by four {\ratsql} layers, and trained on the train split of Spider dataset. The dev set used used for selecting the best model excludes data from the four held-out evaluation groups.

\noindent
{\bf Edit and Fill model}: We utilize a pre-trained \textsc{BART-base} as our conditional text generation model for editing and filling the masked text. The model is fine-tuned using the train split of Spider dataset as described in Section~\ref{sec:schema_translation}

\noindent
{\bf Filtering Model}: We train a binary classifier based on a \textsc{ROBERTA-base} checkpoint on Spider's train split to filter out inconsistent SQL-Text pairs as described in Section~\ref{sec:filter_desc}.

\noindent
{\bf Baselines}: For baseline {\sqltotext} generation models, we consider recently proposed models like {\ltos}~\cite{victorialin2021}, {\gazp}~\cite{grounded-2020-zhang}, and {\snowball}~\cite{shu2021logic}.  All the baselines utilize pre-trained language models like {\bart}~\cite{lewis2020bart} or {\bert}~\cite{devlin2018bert} for translating SQL tokens to natural text in a standard seq-to-seq set-up. The baselines mostly differ in the way of encoding SQL tokens as an input to the language model. In Section~\ref{sec:related_work}, we reviewed the recent {\sqltotext} methods.

\noindent
{\bf {Evaluation Metrics}}
We evaluate the {\texttosql} parsers using the Exact Set Match (EM), and the Exection Accuracy (EX)~\citet{spider2018yu}. The EM metric measures set match for all the SQL clauses and returns $1$ if there is a match across all the clauses. It ignores the DB-values (constants) in the SQL query. The EX metric directly compares the results obtained by executing the predicted query $\hat{\vq}$ and the gold query $\vq$ on the database.

We provide more implementation details including the hyperparameter settings in appendix~\ref{sec:hyperparams}.

\section{Results and Analysis}
We first demonstrate the effectiveness of the synthetic data generated using {\sysname} for fine-tuning {\texttosql} parsers to new schemas.
We compare with the recent methods that utilize {\sqltotext} generation for training-data augmentation (\S~\ref{sec:main_results}). %
We then evaluate the intrinsic quality of the synthetic data generated by different methods in terms of the text diversity and the agreement of the generated text with the ground truth (\S~\ref{sec:quality_and_diversity}). %
We demonstrate that higher text diversity results in better performance of the adapted parsers  (\S~\ref{sec:diversity_importance}).
We then justify the key design choices related to masking of the retrieved text and training of the schema translator module that improves the quality of {\sysname} generated text (\S~\ref{sec:designchoice}). %
Finally, we demonstrate the importance of using an independent binary classifier over cycle-consistency filtering (\S~\ref{sec:filtering}). 

\subsection{Evaluating adapted parsers}
\label{sec:main_results}
\begin{figure}[t]
\centering
\includegraphics[width=\columnwidth]{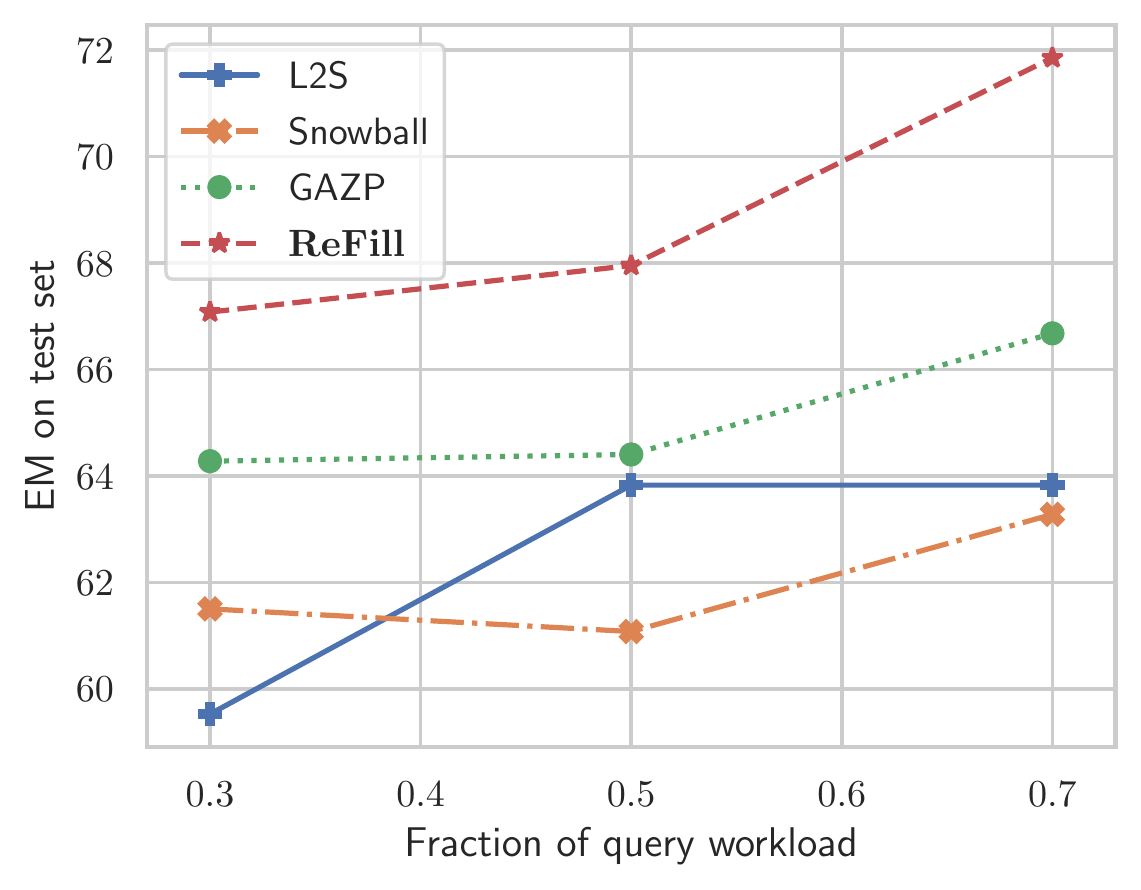}
\caption{Average EM performance of {\texttosql} models on the four groups vs. the size of SQL workload~(\S~\ref{sec:main_results}). Data generated by {\sysname} using 30\% SQL workload yields better performance than the data from the existing best baseline utilizing 70\% workload. }
\label{fig:workload_ablation}
\end{figure}

In Table~\ref{tab:smbop_results}, we
compare the performance of parsers fine-tuned on Text-SQL pairs generated using {\sysname} and other {\sqltotext} generation baselines. We observe that fine-tuning on high-quality and diverse text generated by {\sysname} provides consistent performance gains over the base model across all the database groups. On average, {\sysname} improves the base model by 8.0 EM in comparison to a gain of 2.8 EM by the best baseline (\gazp). We observe that the gains from baseline methods are often small or even negative.   {\sysname} continues to yield positive gains even for smaller workload sizes --- In Figure~\ref{fig:workload_ablation}, we plot the fraction of the total SQL workload used on the x-axis and the EM of the fine-tuned parsers averaged across all the four groups, on the y-axis.
When using the data synthesized by {\sysname}, the performance of the parser improves steadily with an increasing size of the SQL workload. In contrast, the baseline {\sqltotext} generation methods fail to provide significant improvements. Interestingly, the data synthesized by {\sysname} using the 30\% SQL workload leads to better downstream performance of the adapted parsers than any of the baselines utilizing 70\% SQL workload for {\sqltotext} generation. 

\subsection{Quality and Diversity of generated text}
\label{sec:quality_and_diversity}
We explain our gains over existing methods due to the increased quality and diversity of the generated text.
We measure quality using the BLEU score of the set  $S(\vq)$  of generated text for an SQL $\vq$, with the gold text of $\vq$ as reference. To measure diversity we utilize SelfBLEU~\cite{selfbleu} that measures the average BLEU score among text in $S(\vq)$. Lower SelfBLEU implies higher diversity. We evaluate on all the gold SQL-Text pairs available in the Spider's dev set.
In Table~\ref{tab:quality}, we compare the quality and diversity of the text generated using {\sysname} with prior {\sqltotext} generation methods.
For each method we generate 
10 hypotheses per SQL query, and pick the hypothesis with the highest BLEU to report the overall BLEU scores. To allow baselines to generate more diverse text than the standard beam search, we utilize beam-sampling~\cite{fan-etal-2018-hierarchical, holtzman2019curious}.  For {\sysname}, the 10 hypothesis come from using upto 10 retrieved-and-masked templates. 
We observe that our method of masking and refilling the natural text retrieved from existing datasets allows {\sysname} to generate higher quality text (+8.4 BLEU) with naturally high text diversity. %

\begin{table}
\begin{tabular}{l|r|r}
Method   & BLEU $\uparrow$    & 100-SelfBLEU $\uparrow$   \\ 
  & (Quality) & (Diversity) \\
\hline
Gold-Ref & 100    &   68.8 \\ \hline
L2S      &  38.0            &  2.2            \\
GAZP     &  38.8            &  2.0            \\
SnowBall &  40.2            &   2.8           \\
{\sysname} & \textbf{48.6}  &   \textbf{33.8} \\ %

\end{tabular}
\caption{Comparison of quality (BLEU) and diversity (100-SelfBLEU) scores across various {\sqltotext} models including {\sysname} (\S~\ref{sec:quality_and_diversity}). Gold-Ref represents the scores corresponding to the reference text.}
\label{tab:quality}
\end{table}

\subsection{Importance of Text Diversity}
\label{sec:diversity_importance}
\begin{figure}[t]
\centering
\includegraphics[width=\columnwidth]{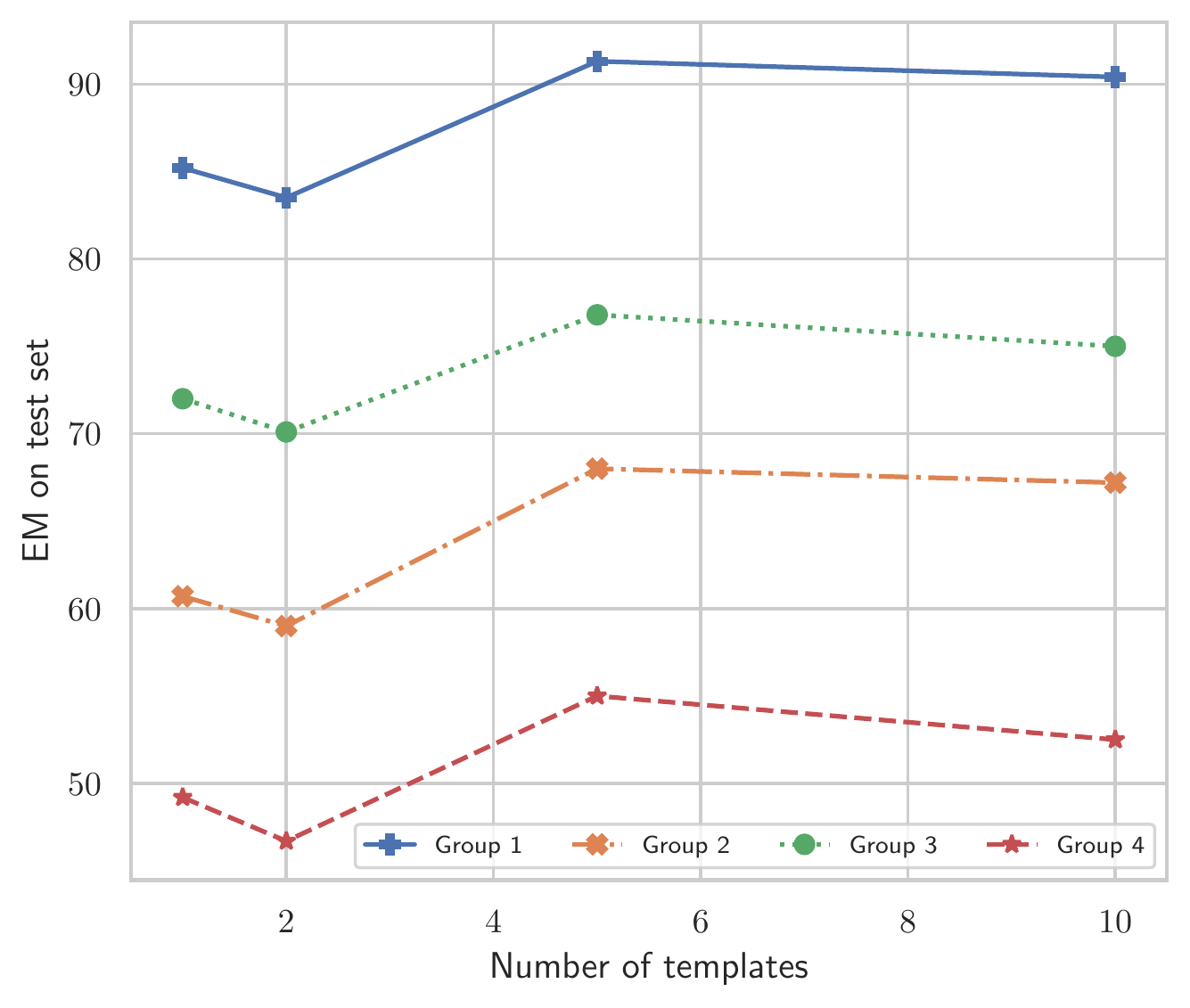}
\caption{Accuracy of fine-tuned parsers Vs. the number of templates per SQL used by {\sysname} (\S~\ref{sec:diversity_importance}).}
\label{fig:diversity}
\end{figure}
Retrieving and editing text from multiple existing examples enables {\sysname} to generate diverse text. In Figure~\ref{fig:diversity}, we show that increased diversity of the generated text leads to improved performance of the fine-tuned parser. We vary the number of retrieved-and-masked templates on the x-axis and plot the performance of the fine-tuned parsers on the y-axis for each group. To maintain the number of synthesized examples the same, the product of beam-samples and the number of retrieved templates is held constant. We observe that fine-tuning the parser on more diverse data generated using 5 retrieved templates per SQL provides consistently superior EM performance across all the four groups than using less diverse data obtained by retrieving just one or two templates per SQL. The consistent drops in EM while increasing the retrieved templates from 5 to 10 is explained by the reduction in text diversity. Using 5 retrieved templates yields a $100-\text{SelfBLEU}$ score of 46.7, while with 10 retrieved templates  $100-\text{SelfBLEU}$ reduces to 33.8. This reduction is due to the inclusion of more similar templates as we increase their number from 5 to 10. Finally, the drop in {\sysname}'s performance with reduced text diversity reconfirms the worse performance of {\sqltotext} baselines reported in Section~\ref{sec:main_results} that do not offer enough text diversity.

\subsection{Design choices of Schema Translator}
\label{sec:designchoice}
\begin{table}[t]
\centering
\begin{tabular}{l|r|r}
     & Na\"ive Train & Robust Train \\ \hline
Schema-Match  & 37.2        & 41.8           \\ \hline
Frequency & 40.2        & 43.8 
\end{tabular}
\caption{Analyzing the impact of design choices related to Schema Translation, by observing BLEU-4 scores of the text generated by {\sysname} (\S~\ref{sec:designchoice}). Frequency based masking and Robust training leads to a higher quality of the generated text.}
\label{tab:design_choice}
\end{table}

In Section~\ref{sec:schema_translation}, we described two important design choices: (1)~Method of masking schema-relevant tokens and (2)~Method of training the Edit-and-Fill model for editing and refilling the masked text. 
We justify these design choices by comparing the quality of the generated text with each combination of these choices in Table~\ref{tab:design_choice}.   Comparing across rows (Schema-Match Vs Frequency), we observe that Frequency based masking results in 2 to 3 point improvements in  BLEU compared to masking by matching schema names. Table~\ref{tab:mask_anecdotes} shows specific examples where the schema-match method fails to mask sufficiently.  In contrast, even though the frequency-based method might over-mask it still suffices for our goal of guiding the text generation model. Comparing across columns (Na\"ive Train Vs. Robust Train) we observe that specifically training the template filling model for being robust to the input templates also improves quality of the generated text by 3.6 to 4.6 points.

\begin{table}[t]
\centering
\begin{tabular}{l|r|r}
                     & EM & EX \\ \hline
{\baseline}           &  45.8  & 35.8   \\ \hline
No Filtering         &  40.8  & 31.7   \\
Cycle Consistent    &  29.2  & 22.5   \\
Filtering Model &  \textbf{48.3}  & \textbf{36.7} 
\end{tabular}
\caption{Using an independent filtering model allows us to retain more useful training examples than cycle consistent filtering, leading to better performance of the fine-tuned {\texttosql} models (\S~\ref{sec:filtering}).}
\label{tab:filtering}
\end{table}
\subsection{Importance of Filtering model}
\label{sec:filtering}
Cycle-consistency based filtering~\cite{grounded-2020-zhang} rejects a synthesized SQL-Text pair $(\vq, \vx)$ if the output SQL $\hat{\vq}$ generated by the base {\texttosql} parser for the input text $\vx$ does not match with the SQL $\vq$. We argue that cycle-consistency based filtering is sub-optimal for two reasons: (i)~{\em Data Redundancy}: Since the {\texttosql} parser is already capable of generating the correct output for the retained examples, fine-tuning on these examples does not offer much improvements. (ii)~{\em~Data~Loss}: If the base {\texttosql} model is weak in parsing text-queries for the target database, a large portion of potentially useful training examples get filtered out due to cycle-inconsistency. In response, we train a Filtering model as described in Section~\ref{sec:filter_desc}. Since our filtering model is independent of the base parser, it retains many useful generated examples that might otherwise be filtered out by cycle-consistency filtering using a weak {\texttosql} parser. In appendix Table~\ref{tab:extra_filtering_examples}, we provide instances of useful training examples that get filtered because of cycle-inconsistency, but are retained by our filtering model. In Table~\ref{tab:filtering}, we compare the base {\texttosql} parser, with models fine-tuned without any filtering, with cycle-consistent filtering, and with using our filtering model. We focus on Group-4 where the base {\texttosql} parser is significantly weaker compared to other groups, and use {\sysname} to synthesize data for the 30\% query workload. %
Not using any filtering, or using cycle-consistent filtering results in worse performance, while applying our filtering model offers significant improvements over the base model.  

\subsection{Experiments on datasets outside Spider}
\label{sec:additional_datasets}
We further validate our method on four single-database datasets outside Spider, namely GeoQuery~\cite{geoquery-original}, Academic~\cite{data-academic}, IMDB and Yelp~\cite{data-sql-imdb-yelp}. %
In Table~\ref{tab:non-spider-results}, we first report the performance of our base {\texttosql} parser %
and observe poor cross-database generalization with an average EM of just 9.7. We adapt the base parser by fine-tuning it on the synthetic datasets generated by {\sysname} and other baselines utilizing 30\% of the available SQL workload. Table~\ref{tab:non-spider-results} compares the EM accuracy of the adapted parsers. We continue to observe that fine-tuning on datasets synthesized by {\sysname} consistently outperforms prior {\sqltotext} based data-augmentation baselines. In appendix Table~\ref{tab:smbop_non_spider_results_full}, we provide results for 50\% and 70\% workload settings. 
\begin{table}[t]
    \centering
    \begin{adjustbox}{width=0.45\textwidth}
    \begin{tabular}{l|r|r|r|r||r}
        Method & Geo & Acad & IMDB & Yelp & Average \\\hline 
        \baseline &  9.8 &  2.8 & 19.3 &  7.2 & 9.7\\
        \ltos     & 27.9 & 14.4 & 24.8 & 19.8 & 21.7\\ 
        \gazp     & 20.8 & 16.0 & 19.3 & 10.8 & 16.7\\ 
        \snowball & 25.6 & 8.8 & 21.1 & 18.9 & 18.6\\
        \sysname (Ours)  & \textbf{30.1} & \textbf{27.6} & \textbf{26.6} & \textbf{29.7} & \textbf{28.5}\\
    \end{tabular}
    \end{adjustbox}
    \caption{Evaluation on datasets outside Spider. We continue to observe that fine-tuning on data synthesized by {\sysname} offers superior EM performance (\S~\ref{sec:additional_datasets}).}
    \label{tab:non-spider-results}
\end{table}

\section{Conclusion}
We presented {\sysname}, a framework for generating diverse and high quality parallel data for adapting existing {\texttosql} parsers to a target database. {\sysname} translates a given SQL workload into a dataset of Text-SQL pairs containing diverse text queries. To achieve higher diversity, {\sysname} retrieves and edits examples from existing datasets and transfers them to a target schema. Through extensive experiments across multiple databases, we demonstrate that fine-tuning {\texttosql} parsers on datasets generated using {\sysname} result in more accurate adaptation to target schemas. %
Even with smaller SQL workloads {\sysname} often outperforms the {\sqltotext} generation baselines utilizing larger SQL workloads. %

\section{Limitations}
This work focuses on synthesizing parallel data containing diverse text queries for adapting pre-trained Text-to-SQL models to new databases. Thus, our current effort toward diverse text query generation using {\sysname} is limited to the Text-to-SQL semantic parsing task. Extending {\sysname} for data-augmentation in other semantic parsing or question-answering tasks is an exciting direction we hope to explore as part of future work. 

Our experimental set-up assumes a small workload of real SQL queries. As per~\citet{BaikJ019}, a small workload of real SQL queries is a reasonable assumption since SQL query logs are often available for existing in-production databases that are to be supported by a Text-to-SQL service. %
Synthesizing realistic SQL-query workloads for newly instantiated databases is a challenging and promising direction but different from the diverse text-query generation aspect of our work.    

\section{Ethical Considerations}
Our goal with {\sysname} is to synthesize parallel data for adapting Text-to-SQL parsers to new schemas. We believe that the real-world deployment of Text-to-SQL or any semantic parser trained on text generated by language models must go through a careful review of any harmful biases. Also, the intended users of any Text-to-SQL service must be made aware that the answers generated by these systems are likely to be incorrect. We do not immediately foresee any serious negative implications of the contributions that we make through this work.

\section{Acknowledgements}
We thank the reviewers from ACL rolling review  for their insightful and quality feedback. We gratefully acknowledge support from IBM Research, specifically the IBM AI Horizon Networks-IIT Bombay initiative. Abhijeet thanks Google for supporting his research with Google PhD Fellowship.

\bibliography{custom}
\bibliographystyle{acl_natbib}

\clearpage %
\appendix

\onecolumn{
\section{Appendix}
\setcounter{table}{0}
\renewcommand{\thetable}{A\arabic{table}}
\subsection{Dataset Details}
\label{sec:grouping_details}
\begin{table}[H]
\centering
\small

\begin{tabular}{l|rrrr|r|r}
Group & \multicolumn{6}{c}{Number of queries by hardness}  \\ 
      \cline{2-7}
      & easy & medium & hard & extra & total examples & unique SQLs\\\hline
      
Group 1                  & 24 & 60 & 25 & 6 & 115 & 62 \\ 
$\bullet$ \texttt{concert\_singer} & 4  & 24 & 13 & 4 & 45 & 26 \\
$\bullet$ \texttt{singer}          & 6  & 18 & 6  & 0 & 30 & 16 \\ 
$\bullet$ \texttt{orchestra}       & 14 & 18 & 6  & 2 & 40 & 20 \\ \hline 

Group 2               & 14 & 58 & 16 & 36 & 124 & 65\\ 
$\bullet$ \texttt{dog\_kennels} & 10 & 36 & 10 & 26 & 82 & 42\\ 
$\bullet$ \texttt{pets\_1}      & 4  & 22 & 6  & 10 & 42 & 23\\ \hline 

Group 3                                  & 46 & 60 & 32 & 26 & 164 & 84\\
$\bullet$ \texttt{students\_transcripts\_tracking} & 26 & 24 & 8  & 20 & 78 & 41\\ 
$\bullet$ \texttt{course\_teach}                   & 8  & 14 & 8  & 0  & 30 & 15\\ 
$\bullet$ \texttt{network\_1}                      & 12 & 22 & 16 & 6  & 56 & 28\\ \hline 

Group 4           & 24 & 46 & 20 & 60 & 120 & 65\\ 
$\bullet$ \texttt{world\_1} & 24 & 46 & 20 & 60 & 120 & 65\\ 
\end{tabular}
\caption{Number of schemas and statistics of query workload for each group. Related schemas were grouped together in order to obtain larger evaluation sets per group.} %
\label{tab:grouping_details}
\end{table}

\begin{table}[H]
    \centering
    \small
\begin{tabular}{l|rrrr|r|r}
Dataset & \multicolumn{6}{c}{Number of queries by hardness}  \\ 
      \cline{2-7}
      & easy & medium & hard & extra & total examples & unique SQLs\\\hline
Geoquery (Geo)    & 224 &  32 & 220 &  77 & 553 & 210 \\
Academic (Acad)   &  20 &  29 &  25 & 107 & 181 & 176\\
IMDB              &  23 &  12 &  48 &  26 & 109 & 75\\
Yelp              &  13 &  29 &  25 &  24 & 111 & 98\\
\end{tabular}
    
    \caption{Statistics of queries in additional (non-spider) datasets. We utilize the pre-processed versions of these datasets provided by~\citet{spider2018yu}.} %
    \label{tab:non_spider_details}
\end{table}

\subsection{Results in low or medium SQL workload setting}
\begin{table}[H]
    \centering
\begin{tabular}{l| r|r | r|r | r|r | r|r || r|r }
  & \multicolumn{2}{c|}{Group 1} & \multicolumn{2}{c|}{Group 2} & \multicolumn{2}{c|}{Group 3} & \multicolumn{2}{c||}{Group 4} & \multicolumn{2}{c}{Average} \\\hline%

Method      &  EM  &  EX  &    EM  &  EX    &  EM  & EX   &   EM   & EX   &    EM   & EX   \\\hline%
\multicolumn{11}{c}{Results with 30\% SQL workload} \\\hline
{\baseline}    & 80.9 & 84.3 &   64.8 & \textbf{67.2}   & 64.0 & 65.9 &   45.8 & 35.8 & 63.8 & 63.3  \\
{\ltos}     & 82.6 & 84.3 &   60.5 & 65.3   & 61.6 & 63.4 &   26.7 & 26.7 &   57.8 & 59.9  \\
{\gazp}     & 83.5 & 84.3 &   61.3 & 64.5   & 66.5 & 67.1 &   45.8 & 37.5 &   64.3 & 63.3  \\
{\snowball} & 80.0 & 83.5 &   59.7 & 63.7   & 67.7 & \textbf{68.3} &   39.2 & 32.5 &   61.6 & 62.0  \\
{\sysname}  & \textbf{86.1} & \textbf{86.1} &   \textbf{65.6} & 65.6   & \textbf{68.3} & 67.1 &   \textbf{48.3} & \textbf{36.7} &   \textbf{67.1} & \textbf{63.8}  \\
\hline\multicolumn{11}{c}{Results with 50\% SQL workload} \\\hline
{\baseline}    & 80.9 & 84.3 &   64.8 & 67.2   & 64.0 & 65.9 &   45.8 & 35.8 &  63.8 & 63.3  \\ %
{\ltos}     & \textbf{89.6} & 88.7 &   66.1 & 68.5   & 57.9 & 58.5 &   41.7 & 35.8 &  63.8 & 62.8  \\
{\gazp}     & 87.8 & 87.0 &   58.9 & 63.7   & 65.9 & \textbf{68.9} &   45.0 & 35.0 &  64.4 & 63.6  \\
{\snowball} & 83.5 & 85.2 &   55.6 & 66.1   & 65.2 & 66.5 &   40.0 & 32.5 &   61.1 & 62.6  \\
{\sysname}  & 88.7 & \textbf{91.3} &   \textbf{67.2} & \textbf{69.7}   & \textbf{70.7} & 67.1 &   \textbf{45.8} & \textbf{38.3} &   \textbf{68.1} & \textbf{66.6}  \\
\hline\multicolumn{11}{c}{Results with 70\% SQL workload} \\\hline
{\baseline}    & 80.9 & 84.3 &   64.8 & 67.2   & 64.0 & 65.9 &   45.8 & 35.8 &   63.8 & 63.3\\
{\ltos}    & \textbf{88.7} & \textbf{87.8} &   61.3 & 62.1   & 62.8 & 61.0 &   42.5 & 35.0 &   63.8 & 61.4\\
{\gazp}     & 85.2 & 85.2 &   58.9 & 66.9   & 70.1 & 60.5 &   52.5 & 40.8 &   66.6 & 63.3\\
{\snowball} & 85.2 & \textbf{87.8} &   59.7 & 60.5   & 64.0 & 65.9 &   44.2 & 38.3 &  63.2 & 63.1\\
{\sysname}   & \textbf{88.7} & 87.0 &   \textbf{69.7} & \textbf{73.8}   & \textbf{73.2} & \textbf{70.1} &   \textbf{55.8} & \textbf{45.0} &   \textbf{71.8} & \textbf{68.9}\\
\end{tabular}

    \caption{Evaluation on four groups of schemas held out from Spider's dev set, for varying sizes of query workload \{30\%, 50\%, 70\%\} used for {\sqltotext} translation.}
    \label{tab:smbop_spider_results}
\end{table}

\begin{table}[H]
\begin{minipage}{0.45\linewidth}
\small
    \centering
    \begin{tabular}{l|r|r|r|r||r}
    Method & Geo & Acad & IMDB & Yelp & Average \\\hline %
    \multicolumn{6}{c}{Results with 30\% SQL workload} \\\hline%
    \baseline &  9.8 &  2.8 & 19.3 &  7.2  &  9.7\\
    \ltos     & 27.9 & 14.4 & 24.8 & 19.8  & 21.7\\
    \gazp     & 20.8 & 16.0 & 19.3 & 10.8  & 16.7\\
    \snowball & 25.6 &  8.8 & 21.1 & 18.9  & 18.6\\
    \sysname  & \textbf{30.1} & \textbf{27.6} & \textbf{26.6} & \textbf{29.7} & \textbf{28.5}\\ 
    
    \hline\multicolumn{6}{c}{Results with 50\% SQL workload} \\\hline%
    \baseline & 9.8 &  2.8 & 19.3 &  7.2  &  9.7\\
    \ltos     & \textbf{33.6} & 20.4 & 25.7 & 18.0 & 24.4\\
    \gazp     & 21.5 & 11.1 & 23.9 & 14.4 & 17.7\\
    \snowball & 26.0 & 24.3 & 18.3 & 27.9 & 24.1\\
    \sysname  & 27.9 & \textbf{37.6} & \textbf{28.4} & \textbf{35.1}  & \textbf{32.2}\\
    
    \hline\multicolumn{6}{c}{Results with 70\% SQL workload} \\\hline%
    \baseline &  9.8 &  2.8 & 19.3 &  7.2 &  9.7\\
    \ltos     & \textbf{33.9} & 19.3 & 29.4 & 23.4 & 26.5 \\ 
    \gazp     & 25.4 & 13.8 & 22.0 & 15.3 & 19.1\\ 
    \snowball & 30.9 & 20.9 & 20.1 & \textbf{35.1} & 26.7\\
    \sysname  & 32.8 & \textbf{37.0} & \textbf{33.0} & \textbf{35.1} & \textbf{34.4} \\
    \end{tabular}
     \caption{EM evaluation on four additional datasets outside Spider, for varying sizes of query workload \{30\%, 50\%, 70\%\} used for {\sqltotext} translation. Since the contents of Acad, IMDB, and Yelp databases were not publicly accessible to us, we are unable to report EX results on these databases. EX results for GeoQuery appear in Table~\ref{tab:smbop_geo_ex}.}
    \label{tab:smbop_non_spider_results_full}
\end{minipage}
\hfill
\begin{minipage}{0.45\linewidth}
    \centering
    \begin{tabular}{l|ccc}
               & \multicolumn{3}{c}{Fraction of SQL workload} \\ 
               \cline{2-4}
        Method & 30\% & 50\%  & 70\% \\\hline
        \baseline & 27.2 & 27.2 & 27.2 \\
        \ltos     & 30.4 & \textbf{35.1} & 37.5 \\
        \gazp     & 20.8 & 22.9 & 29.2 \\
        \snowball & 28.7 & 28.5 & 33.8 \\
        \sysname  & \textbf{33.0} & 34.2 & \textbf{38.9} \\
    \end{tabular}
    \caption{EX accuracy evaluation on GeoQuery dataset, for varying sizes of query workload \{30\%, 50\%, 70\%\}.}
    \label{tab:smbop_geo_ex}
\end{minipage}
\end{table}

\subsection{Examples rejected by cycle-consistency but retained by our filtering model} 
\begin{table}[H]
\small
\centering

\begin{adjustbox}{width=\columnwidth}
\begin{tabular}{p{0.15\columnwidth} | p{0.85\columnwidth}}

Generated text & How many countries are governed by Islamic Emirate?\\
Gold SQL & \small{\texttt{SELECT count(*) FROM country WHERE GovernmentForm = 'Islamic Emirate'}}\\
Predicted SQL & \small{\texttt{SELECT COUNT(*) FROM country WHERE country.code NOT IN (SELECT countrylanguage.countrycode FROM countrylanguage)}}\\\hline 
Generated text & What is the number of languages that are official in Australia?\\
Gold SQL & \small{\texttt{SELECT COUNT(*) FROM country AS T1 JOIN countrylanguage AS T2 ON T1.Code = T2.CountryCode WHERE T1.Name = 'Australia' AND IsOfficial = 'T'}}\\
Predicted SQL & \small{\texttt{SELECT COUNT(*) FROM countrylanguage JOIN country ON countrylanguage.countrycode = country.code WHERE country.name = 'Australia'}}\\\hline 
Generated text & How many countries have both ``Karen" and ``Mandarin Chinese" languages?\\
Gold SQL & \small{\texttt{SELECT COUNT(*) FROM (SELECT T1.Name FROM country AS T1 JOIN countrylanguage AS T2 ON T1.Code = T2.CountryCode WHERE T2.Language = 'Karen' INTERSECT SELECT T1.Name FROM country AS T1 JOIN countrylanguage AS T2 ON T1.Code = T2.CountryCode WHERE T2.Language = 'Mandarin Chinese')}}\\
Predicted SQL & \small{\texttt{SELECT COUNT(*) FROM countrylanguage JOIN country ON countrylanguage.countrycode = country.code WHERE countrylanguage.language = 'Karen'}} \\\hline 
Generated text & Find the language of the country that has the head of state Salahuddin Abdul Aziz Shah Alhaj and is official. \\
Gold SQL & \small{\texttt{SELECT T2.Language FROM country AS T1 JOIN countrylanguage AS T2 ON T1.Code = T2.CountryCode WHERE T1.HeadOfState = 'Salahuddin Abdul Aziz Shah Alhaj' AND T2.IsOfficial = 'T'}} \\
Predicted SQL & \small{\texttt{SELECT countrylanguage.language FROM countrylanguage JOIN country ON countrylanguage.countrycode = country.code WHERE country.headofstate = 'Salahuddin Abdul Aziz Shah Alhaj'}} \\\hline 
Generated text & What are the names of countries with surface area greater than the smallest area of any country in Antarctica?\\
Gold SQL & \small{\texttt{SELECT Name FROM country WHERE SurfaceArea > (SELECT min(SurfaceArea) FROM country WHERE Continent = 'Antarctica')}}\\
Predicted SQL & \small{\texttt{SELECT country.name FROM country WHERE country.surfacearea > (SELECT MAX(country.surfacearea) FROM country WHERE country.continent = 'Antarctica')}}\\
\end{tabular}
\end{adjustbox}
\caption{Consistent SQL-Text pairs rejected by cycle-consistency but retained by our filtering model. Predicted SQL is the output of the {\texttosql} model used for checking cycle consistency, and does not match the gold SQL often due to minor errors.}

\label{tab:extra_filtering_examples}
\end{table}
}

\onecolumn
\renewcommand{\thetable}{A\arabic{table}}
\subsection{Examples of masking}
\input{tables/maskanecdotes}
\FloatBarrier

\twocolumn

\subsection{Utilizing synthetic SQL workload}
    We also experimented with synthetic SQL workloads generated by {\gazp}~\cite{grounded-2020-zhang}. We generate 2000 SQLs for each group and perform data-augmentation using {\sysname}  and other {\sqltotext} baselines. For all the methods, we observe almost consistent drops in model performance after fine-tuning on Text-SQL pairs with synthetically generated SQLs workloads, as presented in the Table~\ref{tab:synthetic_workload}.  A potential reason behind the drops is a mismatch between the distribution of synthetically generated SQLs and the actual distribution of SQLs in the test data~\cite{herzig2019don}. 
    
    \begin{table}[h!]
    \begin{adjustbox}{width=\columnwidth}
    \begin{tabular}{l|r|r|r|r|r|r|r|r}
          & \multicolumn{2}{c|}{Group 1} & \multicolumn{2}{c|}{Group 2} & \multicolumn{2}{c|}{Group 3} & \multicolumn{2}{c}{Group 4} \\ \hline
    Method       & EM           & EX          & EM           & EX          & EM           & EX          & EM           & EX          \\ \hline
    \baseline & \textbf{80.9}         & \textbf{84.3}        & \textbf{64.8}         & \textbf{67.2}        & \textbf{64.0}           & \textbf{65.9}        & 45.8         & 35.8        \\
    {\gazp}   & 76.5         & 76.5        & 51.6         & 58.2        & 55.5         & 59.1        & \textbf{48.3}         & \textbf{40.0}          \\
    {\ltos} & 67.8 & 59.7  &  45.5 & 52.5  &  50.1 & 51.8  &  37.4 & 26.7 \\
    {\sysname} & 76.5 & 75.7  &  52.6 & 58.2  &  57.1 & 58.2  &  45.8 & 33.0    
    \end{tabular}
    \end{adjustbox}
    \caption{Experiments with synthetic SQL workloads generated using the method from {\gazp}. We generate 2000 SQLs for each group. {\sysname} here uses the SQLs synthesized by {\gazp} but uses it's own {\sqltotext} model and filtering model as described in Section~\ref{sec:our_method}. For both the methods we observe that fine-tuning on parallel data generated by translating synthetic workloads hurts model performance.}
    \label{tab:synthetic_workload}
    \end{table}
    
It is important to note that our experimental set-up differs from the original set-up of {\gazp}~\cite{grounded-2020-zhang} or {\ltos}~\cite{victorialin2021}. For example, {\gazp}'s set-up assume that the test-time schemas are known before the base model is trained. Hence, {\gazp} augments the training data used for training the base model with synthetic Text-SQL pairs from the test databases. We instead focus on adapting a parser to databases that were not seen during the train time. Thus, we make no assumptions about the test-time databases while training the base model. In our setting, we find that finetuning on Text-SQL pairs generated using a synthetic SQL workload does not improve the parser performance.

\subsection{Hyperparameters}
\label{sec:hyperparams}
Our Edit and Fill model (139.2M parameters) is based on a pretrained \textsc{BART-base} \cite{lewis2020bart} model. We fine-tune this model for 100 epochs with learning rate of $3\times 10^{-5}$, weight decay of $0.01$ and batch size of 64. The pretrained model is obtained from HuggingFace\footnote{\small{\url{https://huggingface.co/facebook/bart-base}}}. 

The proposed binary classifier (124.6M params) is pretrained \textsc{RoBERTa-base} \cite{liu2020roberta} (obtained from HuggingFace\footnote{\small{\url{https://huggingface.co/roberta-base}}}) fine-tuned for 100 epochs on our data with learning rate $10^{-5}$, weight decay $0.01$ and batch size 16 for 100 epochs. 

For {\smbop} experiments, we use a smaller {\smbop} model with 4 RAT layers and \textsc{RoBERTa-base} \cite{liu2020roberta} encoder as a baseline. The number of parameters in this model is 132.9M. All the adaptation experiments use learning rate of $5\times 10^{-6}$, learning rate of language model of $3\times 10^{-6}$ and batch size of 8. All the models were trained for 100 epochs.

All the experiments were performed on a single NVIDIA RTX 3060 GPU. Training times for the Edit-and-Fill model and binary classifiers were $\approx 4.5$ hrs and $\approx 6.5$ hrs respectively. Each of the fine-tuning experiment took $3-4$ hrs to complete.

\subsection{Cost function for Tree Edit Distance}

\begin{table}[h]
    \centering
    \begin{tabular}{cc|c}
        Group & Value & Cost\\\hline
        Equal & Equal & $0$ \\
        Equal & Unequal & $0.5$ \\ 
        Unequal & Equal & $0$ \\ 
        Unequal & Unequal & $1$ \\
    \end{tabular}
    \caption{Cost function of nodes $n_1$ and $n_2$ based on their groups and value.}
    \label{tab:cost_function}
\end{table}

\begin{table}[h]
    \centering
    \begin{tabular}{p{0.25\linewidth}|p{0.7\linewidth}}
        Group & SQL elements\\\hline
        Aggregation & \texttt{MAX}, \texttt{MIN}, \texttt{AVG}, \texttt{COUNT}, \texttt{SUM} \\
        Order & \texttt{ORDERBY\_ASC}, \texttt{ORDERBY\_DESC} \\
        Boolean & \texttt{OR}, \texttt{AND} \\
        Set & \texttt{UNION}, \texttt{INTERSECT}, \texttt{EXCEPT} \\
        Leaf & \texttt{VAL\_LIST}, \texttt{VALUE}, \texttt{LITERAL}, \texttt{TABLE}\\
        Similarity & \texttt{LIKE}, \texttt{IN}, \texttt{NOT\_IN} \\
        Comparison & $>$, $\geq$, $<$, $\leq$, $=$, $\neq$ \\ 
    \end{tabular}
    \caption{Group definitions for TED calculation.}
    \label{tab:ted_groups}
\end{table}

We use APTED library ~\cite{pawlik2015efficient, pawlik2016tree} to compute TED between 2 parsed SQL trees. For every node in the tree, a group is assigned according to table \ref{tab:ted_groups}. Then the cost for various combinations of node groups and node values is described in table \ref{tab:cost_function}. If either of the nodes does not belong to any of the groups in table \ref{tab:ted_groups}, their groups are considered to be ``unequal" and cost will be assigned based on their values.

\subsection{Examples of TED neighbours}

\begin{figure}[h]
    \centering
    \begin{subfigure}[b]{0.5\textwidth}
    \centering
    \begin{forest}
    [Table(SELECT)[Agg(COUNT)[Value(*)]][Table(FROM)[Predicate(>)[Value(age)][Value(56)]][Table(head)]]]
    \end{forest}
    \caption{\tiny{\texttt{SELECT count(*) FROM head WHERE age > 56}}}
    \end{subfigure}\\
    \begin{subfigure}[b]{0.5\textwidth}
    \centering
    \begin{forest}
    [Table(SELECT)[Agg(COUNT)[Value(*)]][Table(FROM)[Predicate(>)[Value(season)][Value(2007)]][Table(game)]]]
    \end{forest}
    \caption{\tiny{\texttt{SELECT count(*) FROM game WHERE season > 2007}}}
    \end{subfigure}
    \caption{Example of tree pair with TED=$0$}
    \label{fig:zero_ted}
\end{figure}
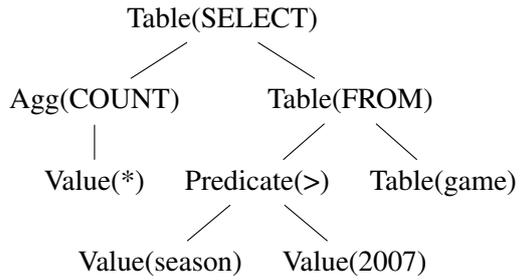

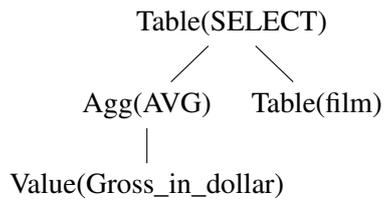
\begin{figure}[h]
    \centering
    \begin{subfigure}[b]{0.5\textwidth}
    \centering
    \begin{forest}
    [Table(SELECT)[Agg(COUNT)[Value(*)]][Table(county\_public\_safety)]]
    \end{forest}
    \caption{\tiny{\texttt{SELECT count(*) FROM county\_public\_safety}}}
    \end{subfigure}\\
    \begin{subfigure}[b]{0.5\textwidth}
    \centering
    \begin{forest}
    [Table(SELECT)[Agg(AVG)[Value(Gross\_in\_dollar)]][Table(film)]]
    \end{forest}
    \caption{\tiny{\texttt{SELECT avg(Gross\_in\_dollar) FROM film}}}
    \end{subfigure}
    \caption{Example of tree pair with non-zero(=0.125) TED} %
    \label{fig:non_zero_ted}
\end{figure}

\end{document}

%% file: tables/main_results.tex
\begin{table*}[h!]
\centering
\begin{adjustbox}{width=\textwidth}
\begin{tabular}{l| r|r | r|r | r|r | r|r || r|r }
  & \multicolumn{2}{c|}{Group 1} & \multicolumn{2}{c|}{Group 2} & \multicolumn{2}{c|}{Group 3} & \multicolumn{2}{c||}{Group 4} & \multicolumn{2}{c}{Average} \\ \hline
Method      &  EM  &  EX  &    EM  &  EX    &  EM  & EX   &   EM   & EX   &    EM   & EX   \\ \hline
{\baseline}    & 80.9 & 84.3 &   64.8 & 67.2   & 64.0 & 65.9 &   45.8 & 35.8 &   63.8 & 63.3\\
{\ltos}~\cite{victorialin2021}     & \textbf{88.7} & \textbf{87.8} &   61.3 & 62.1   & 62.8 & 61.0 &   42.5 & 35.0 &   63.8 & 61.4\\
{\gazp}~\cite{grounded-2020-zhang}     & 85.2 & 85.2 &   58.9 & 66.9   & 70.1 & 60.5 &   52.5 & 40.8 &   66.6 & 63.3\\
{\snowball}~\cite{shu2021logic} & 85.2 & \textbf{87.8} &   59.7 & 60.5   & 64.0 & 65.9 &   44.2 & 38.3 &  63.2 & 63.1\\
{\sysname} (Ours)  & \textbf{88.7} & 87.0 &   \textbf{69.7} & \textbf{73.8}   & \textbf{73.2} & \textbf{70.1} &   \textbf{55.8} & \textbf{45.0} &   \textbf{71.8} & \textbf{68.9}\\
\end{tabular}
\end{adjustbox}
\caption{Results for finetuning a base semantic parser ({\smbop}) on Text-SQL pairs generated using various {\sqltotext} baselines and {\sysname} (\S~\ref{sec:main_results}). {\sysname} provides consistent gains over the base model across all the database groups, while gains from other methods are often negative or small.}
\label{tab:smbop_results}
\end{table*}

%% file: tables/maskanecdotes.tex
\begin{table}[H]
\centering
\begin{adjustbox}{width=\columnwidth}
\begin{tabular}{p{0.15\linewidth} | p{0.85\linewidth}}

SQL                   & \small{\texttt{SELECT T1.template\_type\_code ,  count(*) FROM Templates AS T1 JOIN Documents AS T2 ON T1.template\_id  =  T2.template\_id GROUP BY T1.template\_type\_code}} \;\; [\textbf{Schema Name}: {\color{blue}{\tt Document Template Management}}]      \\
Reference             & Show all template type codes and the number of documents using each type.                                                                                                                       \\ \hline
Retrieved SQL         & \small{\texttt{T1.FacID ,  count(*) FROM Faculty AS T1 JOIN Student AS T2 ON T1.FacID  =  T2.advisor GROUP BY T1.FacID}}  \;\; [\textbf{Schema Name}: {\color{red}{\tt Faculty Student Activity}}] \\
Retrieved Text        & Show the faculty id of each faculty member, along with the number of students he or she advises.                                                                                                                            \\ \hline
Sch-match Mask       & Show the {\mask} of each {\mask} {\color{red}member} , along with the number of {\mask} he or she {\color{red}advises} .                                         \\
Filled Text & Show the {\color{blue}type code} of each {\color{blue}template} {\color{red}member}, along with the number of {\color{blue}documents} he or she {\color{red}advises}.                                                                                                                          \\ \hline
Freq Mask         & Show the {\mask} of each {\mask} , {\mask} with the number of {\mask} he or she {\mask} .                       \\
Filled Text  & Show the {\color{blue}code} of each {\color{blue}template type}, together with the number of {\color{blue}documents} corresponding to it. \\ \hline \hline

SQL                   & \small{\texttt{SELECT T2.name ,  T2.capacity FROM concert AS T1 JOIN stadium AS T2 ON T1.stadium\_id  =  T2.stadium\_id WHERE T1.year  \textgreater{}=  2014 GROUP BY T2.stadium\_id ORDER BY count(*) DESC LIMIT 1}}  \;\; [\textbf{Schema Name}: {\color{blue}{\tt Concert Singer}}]     \\
Reference             & Show the stadium name and capacity with most number of concerts in year 2014 or after.                                                                                                                       \\ \hline
Retrieved SQL         & \small{\texttt{SELECT T2.name ,  T1.team\_id\_winner FROM postseason AS T1 JOIN team AS T2 ON T1.team\_id\_winner  =  T2.team\_id\_br WHERE T1.year  =  2008 GROUP BY T1.team\_id\_winner ORDER BY count(*) DESC LIMIT 1}} \;\; [\textbf{Schema Name}: {\color{red}{\tt Baseball 1}}]\\
Retrieved Text        & What are the name and id of the team with the most victories in 2008 postseason?                                                                                                                             \\ \hline
Sch-match Mask       & What are the {\mask} and {\mask} of the {\mask} with the most {\color{red} victories} in {\mask}                                         \\
Filled Text & What are the {\color{blue}name} and {\color{blue}capacity} of the {\color{blue}stadium} with the most {\color{red} victories} in year {\color{blue}2014}?                                                                                                                          \\ \hline
Freq Mask         & What are the {\mask} and {\mask} of the {\mask} with the most {\mask} in {\mask}                       \\
Filled Text  & What are the {\color{blue}name} and {\color{blue}capacity} of the {\color{blue}stadium} with the most {\color{blue}concerts} in {\color{blue}2014}?

\end{tabular}
\end{adjustbox}
\caption{Masking the text based on string matches Vs. our method of frequency based masking. Schema-relevant words like {\color{red} `victories', `members', `advises'} that do not have a sufficient string match with any of the table or column names of their schema, get left out when using string-match based matches. Thus failing to mask the words in the original schema might lead to copying of the word in the target schema, thus making the generated text semantically inconsistent. Words in {\color{blue} blue} are schema relevant words for the target database and should appear in the generated output.} 
\label{tab:mask_anecdotes}
\end{table}